\documentclass{article}
\usepackage{arxiv}
\usepackage[utf8]{inputenc} 
\usepackage[T1]{fontenc}    
\usepackage{hyperref}       
\usepackage{url}            
\usepackage{booktabs}       
\usepackage{amsfonts}       
\usepackage{nicefrac}       
\usepackage{microtype}      
\usepackage{lipsum}		
\usepackage{graphicx}
\usepackage[numbers]{natbib}
\usepackage{doi}
\usepackage{algorithm}
\usepackage{algpseudocode}
\usepackage{amsfonts}       
\usepackage{amsmath}
\usepackage{bm}
\usepackage{subcaption}

\title{Self-Supervised Autoencoder Network for Robust Heart Rate Extraction from Noisy Photoplethysmogram: Applying Blind Source Separation to Biosignal Analysis}

\author{
  Matthew B. Webster\textsuperscript{1}\thanks{Co-first authors.} \quad
  Dongheon Lee\textsuperscript{2,3}\footnotemark[1] \quad
  Joonnyong Lee\textsuperscript{1} \\
  \\
  \textsuperscript{1}Mellowing Factory Co. Ltd., Seoul, Republic of Korea \\
  \textsuperscript{2}Chungnam National University Hospital, Daejeon, Republic of Korea \\
  \textsuperscript{3}Department of Radiology, Seoul National University College of Medicine, Seoul, Republic of Korea
}
\renewcommand{\shorttitle}{Self-Supervised Autoencoder Network for Robust Heart Rate Extraction}

\hypersetup{
pdftitle={Deep Blind Source Separation for Single-Channel Cardio-Mechanical Signals},
pdfsubject={arxiv},
pdfauthor={Matthew B. ~Webster, Joonnyong ~Lee},
pdfkeywords={Deep learning, Signals Processing, Blind Source Separation},
}

\begin{document}
\maketitle

\begin{abstract}
Biosignals can be viewed as mixtures measuring particular physiological events, and blind source separation (BSS) aims to extract underlying source signals from mixtures. This paper proposes a self-supervised multi-encoder autoencoder (MEAE) to separate heartbeat-related source signals from photoplethysmogram (PPG), enhancing heart rate (HR) detection in noisy PPG data. The MEAE is trained on PPG signals from a large open polysomnography database without any pre-processing or data selection. The trained network is then applied to a noisy PPG dataset collected during the daily activities of nine subjects. The extracted heartbeat-related source signal significantly improves HR detection as compared to the original PPG. The absence of pre-processing and the self-supervised nature of the proposed method, combined with its strong performance, highlight the potential of MEAE for BSS in biosignal analysis.

\end{abstract}


\section{Introduction}
Autoencoders (AEs), neural networks designed to reconstruct input data at the output layer, have proven effective in biosignal processing, including electroencephalogram (EEG) feature extraction \citep{LiJunhua2015}, electrocardiogram (ECG) classification \citep{LuoKan2017}, and biometric data compression \citep{TestaDavid2015}. Despite this, their application in biosignal denoising remains limited \citep{8356704, LeeJoonnyong2018}. In a prior study \citep{WebsterMatthew2024}, we demonstrated that a multi-encoder autoencoder (MEAE) architecture with specialized loss functions could be trained in a self-supervised manner to extract source signals from an original mixed signal, in a process termed blind source separation (BSS). Using this approach, we successfully extracted breathing rate from photoplethysmogram (PPG) and ECG signals, showing results comparable to breathing rates derived from thoracic movement signals measured during sleep. This method’s ability to extract meaningful features in a self-supervised way highlights its potential for biosignal processing, particularly for identifying heart-related source signals from noisy data.

PPG, which measures peripheral pulse via light transmission or reflection in capillaries, is extensively used in healthcare for applications ranging from pulse oximetry in intensive care to heart rate (HR) monitoring in wearable devices \citep{PhanDung2015, TisonGeoffrey2018, TremperKevin1989}. Beyond HR measurement, PPG offers potential for extracting vital cardiovascular information, with applications in cardiac function analysis \citep{AllenJohn2007, ChanPakHei2016, LeeJinseok2013} and blood pressure estimation \citep{WChen2000, LeeJoonnyong2018June, MONTEMORENO2011127, Teng2003, WangLudi2018}. However, PPG is highly susceptible to noise, necessitating robust preprocessing to mitigate both stationary noise (e.g., powerline interference) and non-stationary noise from motion or changes in skin contact. Motion artifacts, in particular, can lead to the reduction of accuracy in extracting cardiovascular features \citep{ElgendiMohamed2012}. For example, studies have shown significant errors in smartwatch HR measurements under motion artifacts compared to ECG references \citep{jpm7020003}. While such inaccuracies may be acceptable for consumer devices, they present challenges for medical-grade applications requiring pulse wave analysis.

Efforts to improve PPG signal-to-noise ratio (SNR) have explored various denoising techniques, including frequency-based filtering \citep{5415601, 4717272, 6905737}, adaptive filtering \citep{7289363, 6111474}, and wavelet-based methods \citep{6908199, 1302650, 5605443}. Each of these approaches has limitations: frequency analysis struggles with non-stationary noise, adaptive filters rely on noise reference signals, and wavelet methods depend on the correspondence between the chosen wavelet and the target signal, which may vary with physiological changes \citep{1302650}. Furthermore, phase shifts introduced by these methods can disrupt time-domain features \citep{FOO200693}. Recently, neural networks (NNs) have gained traction for PPG feature extraction, though their performance has been largely validated only in controlled settings, raising concerns about real-world applicability \citep{Johansson2003-vy, 6549729, 8356704, Terry2004}.

In 2018, we introduced a bidirectional recurrent autoencoder for denoising and extracting HR from highly noisy PPG measured during daily activities, demonstrating its superiority over conventional methods such as wavelet and bandpass filtering \citep{LeeJoonnyong2018}. However, this approach required artificially augmenting the input PPG with simulated noise, which depended on prior knowledge of noise characteristics. This limitation restricted the method’s ability to generalize to other noise types beyond the augmented noise.

This study presents a BSS approach for extracting HR from noisy PPG signals. The method is briefly introduced with reference to prior work, followed by a summary of the datasets used for training the network and for validating HR extraction from noisy PPG. The proposed method’s performance is assessed by comparing extracted HR to simultaneously measured ECG HR. Finally, the implications of BSS for biosignal processing and potential directions for future research are discussed.

\section{Methods}
\subsection{Blind Source Separation Background}
BSS techniques allow for the recovery of source signals from a set of observations without prior knowledge of the source signals or the mixing system. The mixing system is an unknown function that takes in the source signals as an input and produces a set of observations. Typically, a BSS technique attempts to find an approximate inverse of the mixing system that maps observations to the source signals. BSS problems can be classified into three types of mixing scenarios: determined, overdetermined and underdetermined mixtures \citep{bookcommonpierre2010}. A determined mixture is one such that the number of sources is equal to the number of observations. For an overdetermined mixture, the number of observations is greater than the number of sources. Lastly, underdetermined mixtures are those whereby the number of observations is less than the number of sources, and are the most challenging type of mixture as the number of solutions may be infinite without good priors \citep{HYVARINEN1999429}.

BSS techniques often require strong priors in order to be effective in recovering source signals. For example, non-negative matrix factorization (NMF) makes an assumption of linear mixing and non-negative sources \citep{Lee1999}. FastICA also assumes linear mixing and is based on maximizing the statistical independence of sources using non-Gaussianity \citep{NIPS1996_dfd7468a}. In biosignals processing, often various combinations of filtering and preprocessing steps are required for BSS techniques to produce efficacious results due to signal complexity \citep{6144719, s20113238}. Recently, BSS techniques based on deep learning approaches fare well in certain circumstances where a large amount of training data is available, but good priors are not known \citep{WebsterMatthew2024, Brakel2017LearningIF}. Deep learning approaches can be effective due to their ability to learn features and complex relationships from a large distribution, relaxing the need for strong priors. Additionally, deep learning models have the capability to model highly complex non-linear mixing systems.

\subsection{Multi-Encoder Autoencoder for Self-Supervised BSS}
\begin{figure}[htbp]
  \centering
  \includegraphics[width=\linewidth, trim=30 500 40 20, clip]{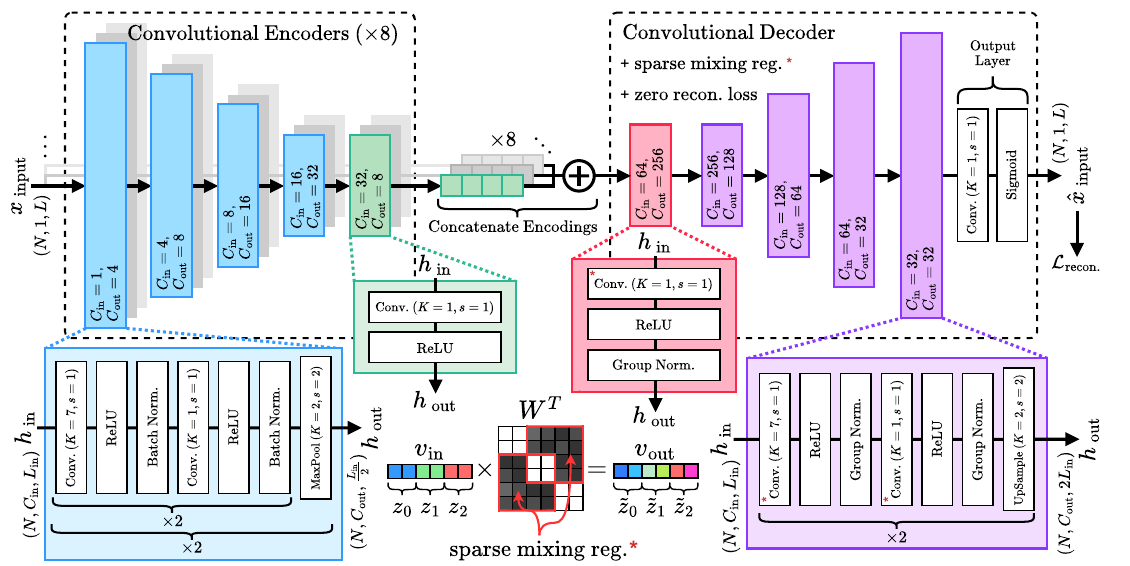}
  \caption{Diagram outlining the architecture of the proposed multi-encoder autoencoder used in this study to extract heart-related source signal from photoplethysmogram. Details of the network and the implementation for training can be found in our GitHub repository.}
  \label{fig:training}
\end{figure}
\begin{figure}[htbp]
  \centering
  \includegraphics[width=0.6\linewidth, trim=20 600 40 20, clip]{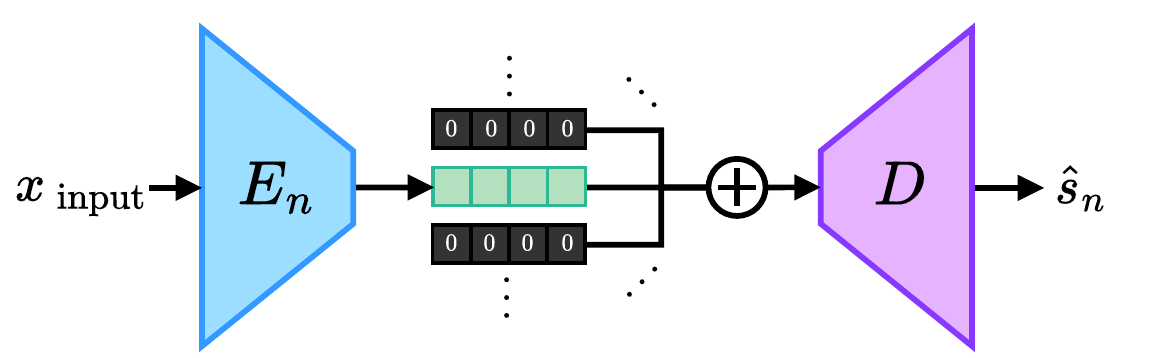}
  \caption{Diagram showing the inference step for the proposed method. An encoder is selected, and the output from all other encoders are masked with zeros. Then the encodings are passed to the decoder, yielding a source prediction corresponding to the selected encoder. Details of the network and the implementation for training can be found in our GitHub repository. $\hat{S}_n$ is the nth predicted source corresponding to the nth encoder, $E_n$.}
  \label{fig:inference}
\end{figure}

The multi-encoder autoencoder (MEAE) architecture introduced in our prior work utilizes a fully convolutional architecture with multiple encoders that all take the same input signal. All of the outputs of the multiple encoders are concatenated along the channel dimension before being passed into the decoder for prediction of the input reconstruction during training as shown in \autoref{fig:training}. Each encoder will learn to specialize in representing a specific source signal, thus to predict any single source during inference (shown in \autoref{fig:inference}) we can mask out all other encoder outputs (i.e. replace the other encodings with all-zero vectors) before decoding. The training process is fully self-supervised meaning that no examples of the sources are provided as a part of the loss. Instead, a reconstruction loss between the input and output is applied along with two additional losses designed specifically for blind source separation. The first loss is the sparse mixing loss which applies an additional weight decay over the weights that connect the separate encodings throughout the decoder. The sparse mixing loss makes the assumption that the mixing system is less complex than the signals themselves and that the sources interact in a sparse manner. The second loss is the zero reconstruction loss. This loss is calculated by first passing all zero vectors into the decoder and then minimizing the output. This loss adds a constraint to the decoder that prevents masked encodings from contributing to the output during source inference. While both of these losses are applied only to the decoder, the encoders still learn source separation by means of the input reconstruction loss with the guidance of the constrained decoder. For specific details about the implementation, see our public repository.
\subsection{MEAE Training Using a Large Polysomnography Dataset}
The Multi-Ethnic Study of Atherosclerosis (MESA) \citep{chen2015racial, Zhang2018TheNS} is a research study involving more than 6000 subjects in the US. The MESA sleep study is a set of polysomnography (PSG) recordings from 2056 patients which consists of simultaneously measured ECG, PPG (recorded at 256Hz), and other biosignals.

Prior to the training process, MESA PPG recordings were resampled to 125Hz to match the sampling rate of the noisy PPG dataset, and each recording was divided into 48 second segments. If the recording duration wasn’t a multiple of 48 seconds, left over seconds were left out at the end of the recording, and segments that had the same minimum and maximum values were removed. The segments were scaled between 0 and 1 using min-max scaling, and padded on both ends by 72 zeros to make the segment length equal to 6144 samples.

The training steps are summarized as followed:
\begin{algorithm}
\caption{Training procedure}
    \begin{algorithmic}[1]
        \State $\bm{E}_{\bm{\theta}}$ is comprised of $N$ encoder networks $E_{\theta_n}^n$ parameterized by $\bm{\theta} = \left[\theta_{0}\ \theta_{1} \ \ldots \ \theta_{N}\right]$.
        \State $D_{\phi}$ is the decoder network parameterized by $\phi$.
        \State $x$ is a batch of samples in the dataset $X$.
        \State Choose weight decay term for off diagonal blocks, $\alpha$.
        \For{$x$ in $X$}
            \State $\bm{z} \gets \bm{E}_{\theta}(x)$ \Comment{Get encodings, $\bm{z} = \left[z^{0}\ z^{1}\ \ldots \ z^{N}\right]$.}
            \State $Z \gets z^{0}\oplus z^{1}\oplus \ldots \oplus z^{N}$ \Comment{Concatenate encodings along channel dimension.}
            \State $\hat{x} \gets D_{\phi}(Z)$ \Comment{Get reconstruction.}
            \State $\mathcal{L}_{\text{recon.}} \gets \text{BCE}(x, \hat{x})$
            \For{$z^n$ in $\bm{z}$} \Comment{Calculate $\mathcal{L}_{\text{z}}$}
                \State $\mathcal{L}_{\text{z}} \gets \mathcal{L}_{\text{z}} + \frac{1}{N h} \|z^n\|_2^2$ \Comment{Apply L2 regularization to encodings.}
            \EndFor
            \For{$W$ in $\phi$}
                \For{$B_{i,j}$ in $W$} \Comment{$B_{i,j}$ are the weight groups in some weight $W$.}
                    \If{$i \neq j$}
                        \State $\mathcal{L}_{\text{mixing}} \gets \mathcal{L}_{\text{mixing}} + \alpha \|B_{i,j}\|_1$ 
                    \EndIf
                \EndFor
            \EndFor
            \State $Z_\text{zero} \gets \left[0\ 0\ 0\ \ldots \ 0\right]$ \Comment{$Z_\text{zero}$ has same shape as $Z$.}
            \State $\mathcal{L}_{\text{zero recon.}} \gets \text{BCE}(0, D_{\phi}(Z_\text{zero}))$
            \State $\mathcal{L}_{\text{total}} = \mathcal{L}_{\text{recon.}} + \lambda_{\text{mixing}} \mathcal{L}_{\text{mixing}} + \lambda_{\text{zero recon.}} \mathcal{L}_{\text{zero recon.}} + \lambda_{\text{z}} \mathcal{L}_{\text{z}}$
            \State Update $\theta$ and $\phi$ by minimizing $\mathcal{L}_{\text{total}}$ via gradient descent.
        \EndFor
    \end{algorithmic}
\label{algo: train}
\end{algorithm}
\autoref{algo: train} outlines the basic training loop for the proposed method. In lines 1-2, we construct the architecture, consisting of $N$ encoders $\bm{E}{\bm{\theta}}$ and a single decoder $D{\phi}$, parameterized by $\theta$ and $\phi$, respectively. In line 4, the decay rate for the off-diagonal blocks, $\alpha$, in the sparse mixing loss is selected.

From lines 5-24, the training loop iterates over batches of samples $x$ from the dataset. In line 6, encodings $\bm{z}$ are obtained by forwarding the batch through each encoder $E_{\theta_n}^n$. Lines 7-9 concatenate these encodings along the channel dimension to form a combined representation, which is then passed through the decoder to produce reconstructions $\hat{x}$. Binary cross-entropy loss is computed between $\hat{x}$ and the original inputs $x$. In lines 10-12, for each encoding $z^n$ in $\bm{z}$, the L2 norm is computed and averaged. Lines 13-19 compute the sparse mixing loss: for each block $B_{i,j}$, such that $i\neq j,$ in each decoder weight layer $W$, the L1 norm is calculated and scaled by $\alpha$, the weight decay strength. In lines 20-21, a second forward pass through the decoder is performed using an all-zero matrix $Z_{\text{zero}}$ of the same shape as the concatenated encodings $Z$, and binary cross-entropy is computed between the output and an all-zero target of the same shape as the input. Finally, in lines 22-23, the individual loss terms are summed to obtain the total loss $\mathcal{L}_{\text{total}}$, which is then minimized via gradient descent. This training loop is repeated until convergence or acceptable performance is achieved.

In this study, we employed eight encoders. During training each of the encoders produced an encoding which were all concatenated to be decoded into the reconstruction signal. During inference a source signal was produced by masking out all but one encoding with all zero matrices before decoding. Specific details including the model hyperparameters and training parameters can be found in our GitHub repository.

\subsection{Validating Noisy PPG Heart Rate Extraction}
To measure noisy PPG simultaneously with reference ECG during daily routines, 9 healthy males (26.9 $\pm$ 1.4 years old) were fitted with a custom chest-worn device for 24 hours. Specific details on the recording hardware can be found in our previous study \citep{LeeJoonnyong2018}. 4 to 6 minutes of PPG and ECG were recorded at 125Hz every 2 hours over 24 hours. A total of 108 recordings representing were used to test the performance of the MEAE in isolating HR from noisy PPG. All subjects provided informed consent and the study was approved by the Institutional Review Board of Seoul National University Hospital.

The PPG recordings were divided into 48 second segments, min-max scaled, and padded prior to inputting to the optimized MEAE. By visual inspection, the most ideal source signal was chosen to represent the HR of the original PPG. The inference steps to extract the encodings are as followed:
\begin{algorithm}
\caption{Inference procedure}
    \begin{algorithmic}[1]
        \State $\bm{E}_{\bm{\theta}}$ is comprised of $N$ encoder networks $E_{\theta_n}^n$ parameterized by $\bm{\theta} = \left[\theta_{0}\ \theta_{1} \ \ldots \ \theta_{N}\right]$.
        \State $D_{\phi}$ is the decoder network parameterized by $\phi$.
        \State $x$ is a sample or a batch of samples in the dataset $X$.
        \State $\bm{z} \gets \bm{E}_{\theta}(x)$ \Comment{Get encodings, $\bm{z} = \left[z^{0}\ z^{1}\ \ldots \ z^{N}\right]$.}
        \State Choose encoding $z^n$ to decode.
        \State $Z^n = \bm{0} \oplus \ldots \oplus z^{n}  \oplus \ldots \oplus \bm{0}$ \Comment{$\bm{0}$ is an all-zero matrix with the same shape as $z^n$}
        \State $\hat{s}^n = D_{\phi}(Z^n)$ \Comment{Get source estimation.}
    \end{algorithmic}
\label{algo: inference}
\end{algorithm}

\autoref{algo: inference} outlines the general inference procedure for generating source estimates using the proposed method. In line 4, the set of encodings $\bm{z}$ is produced for a given input or batch of inputs. In lines 5-6, a single encoding $z^{n}$, the active encoding, is retained for decoding, while all other encodings are replaced with zero matrices $\bm{0}$ matching the original shapes. These encodings are then concatenated along the channel dimension, preserving the position of the active encoding. In line 7, the combined encoding is passed through the decoder $D_{\phi}$ to generate the source estimate $\hat{s}^n$ corresponding to $z^n$.

In order to extract HR from the reference ECG, Two Moving Average Detector from ECG Detectors python library \citep{py-ecg-detectors} was used. For the extraction of HR from the original PPG signal, the point of maximum slope was detected for every cardiac cycle between subsequent ECG R-peaks. For the HR detection in the source signal, the maximum value for the source signal was detected between each subsequent R-peaks \autoref{fig:rpeak}.
\begin{figure}[htbp]
  \centering
  \includegraphics[width=0.55\linewidth, trim=10 440 10 15, clip]{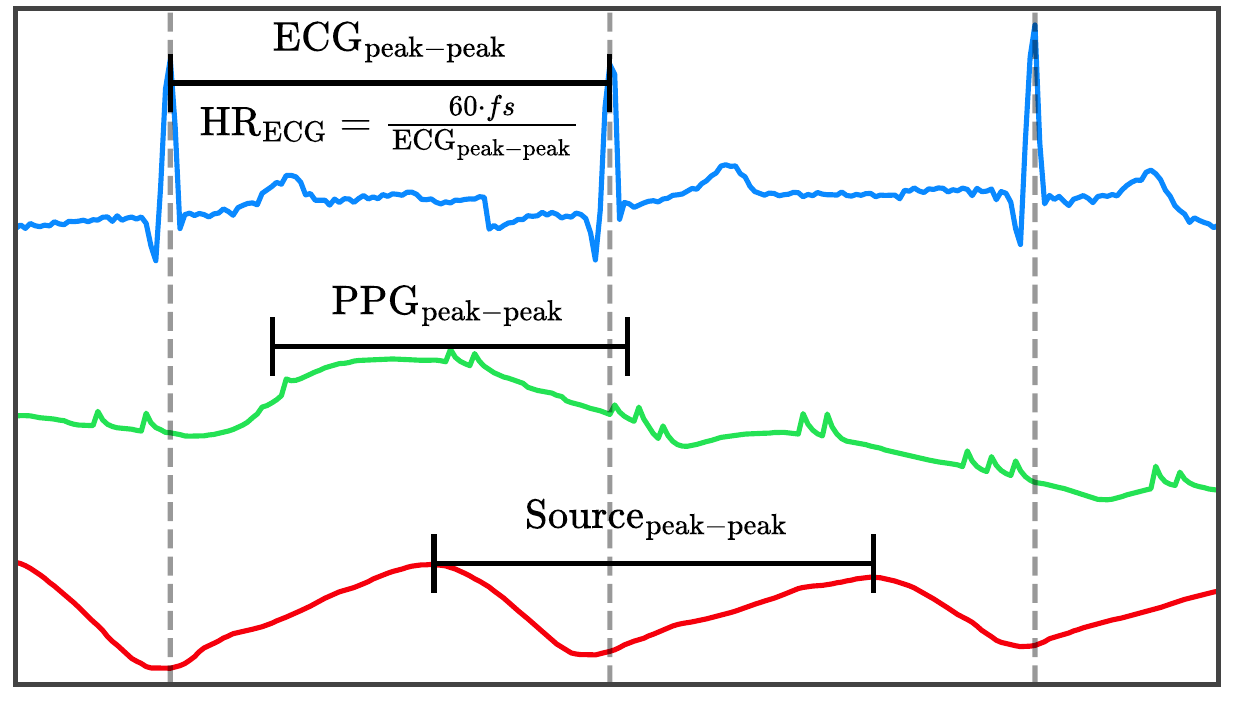}
  \caption{Detection of heart rate from electrocardiogram (top), original photoplethysmogram (middle), and the first source signal (bottom) generated from the optimized multi-encoder autoencoder network. The detected peaks corresponding to each signals are used to generate beat-by-beat heart rates for each cardiac cycle.
}
  \label{fig:rpeak}
\end{figure}

The HR values were detected in 48-second segments and then combined to form beat-by-beat HR arrays for ECG, PPG, and the source signal from the selected encoder for each recording. Median smoothing with window size of 5 cardiac cycles was applied to the detected HR in order to smooth out occasional undetected beats. To compare the extracted HR from the original PPG signal, the source signal, and the reference ECG signal, root mean squared error (RMSE) and Pearson correlation values were calculated for each of the 108 recordings. The results from the proposed method are compared to traditional BSS methods including Independent Component Analysis (ICA) and Non-negative Matrix Factorization (NMF). As the PPG recordings are one dimensional and both the ICA and the NMF requires as many observations as components, 8 pseudo-copies were made by shifting the recording 1 sample per copy, and the best performing components for each method in terms of HR error were selected for comparison. Furthermore, comparison is only made to the results of supervised bidirectional recurrent denoising autoencoder (BRDAE) method in our previous study, and alternative methods in that study are omitted here as the proposed method outperformed the BRDAE, which significantly outperformed other methods including bandpass filtering and wavelet denoising.

\section{Results}
The source signals generated from the network at 45th epoch can be seen in \autoref{fig:rpeak}. The optimized model weights can be found in our GitHub repository and various figures for different forms of PPG from both datasets can be found on Supplementary File 1.
\begin{figure}[htbp]
  \centering
  \begin{subfigure}[b]{0.495\linewidth}
    \centering
    \includegraphics[width=\linewidth]{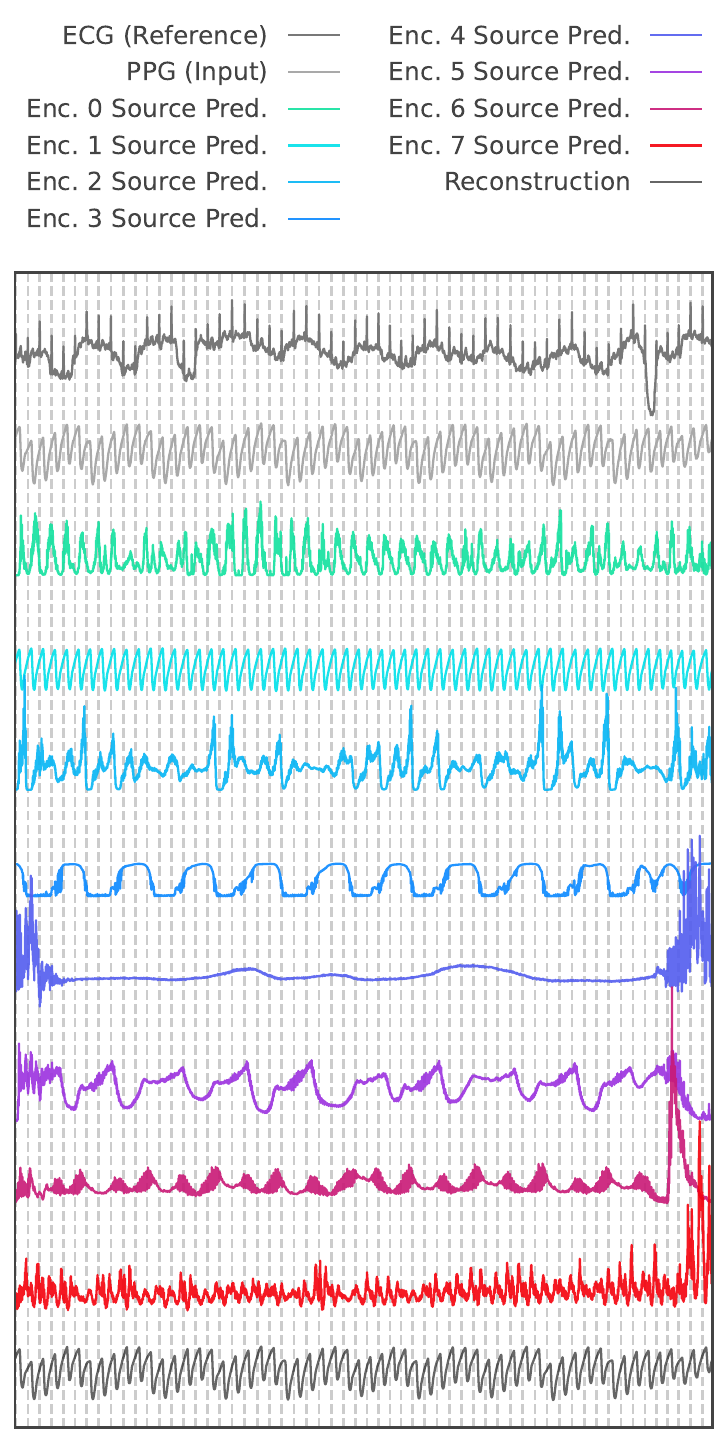}
    \caption{MESA dataset.}
    \label{fig:figure4a}
  \end{subfigure}
  \hfill
  \begin{subfigure}[b]{0.495\linewidth}
    \centering
    \includegraphics[width=\linewidth]{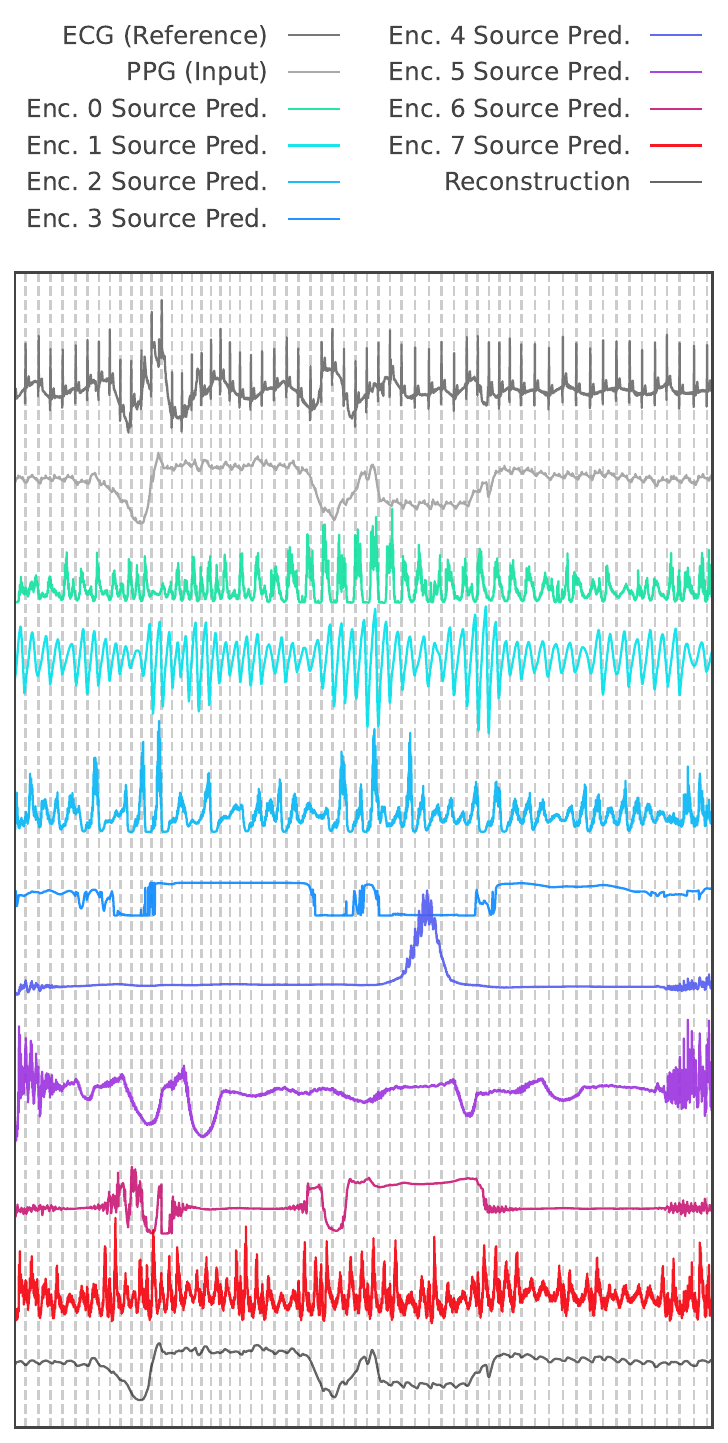}
    \caption{Noisy PPG dataset.}
    \label{fig:figure4b}
  \end{subfigure}
  \caption{Electrocardiogram (top row), photoplethysmogram (2nd row), source signals (3rd–10th rows) from the optimized multi-encoder autoencoder, and PPG reconstruction (bottom row). 48-second samples (clipped on both sides to about 46 seconds to remove noisy edges) from the MESA dataset (a) and the noisy PPG dataset (b) are shown. Dashed vertical lines indicate heartbeat peaks detected from the ECG.}
  \label{fig:rpeak}
\end{figure}

\begin{figure}[htbp]
  \centering
  \begin{subfigure}[b]{0.495\linewidth}
    \centering
    \includegraphics[width=\linewidth]{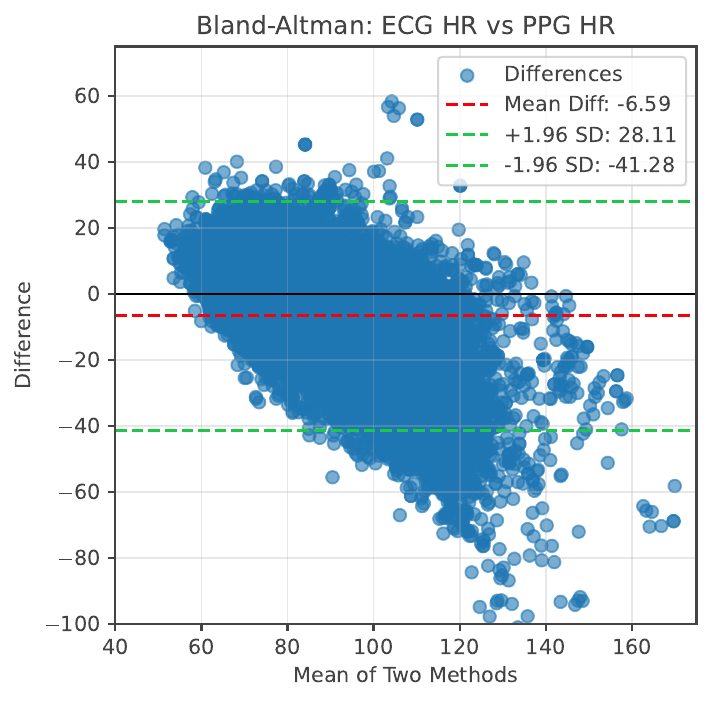}
    \caption{}
    \label{fig:figure5a}
  \end{subfigure}
  \hfill
  \begin{subfigure}[b]{0.495\linewidth}
    \centering
    \includegraphics[width=\linewidth]{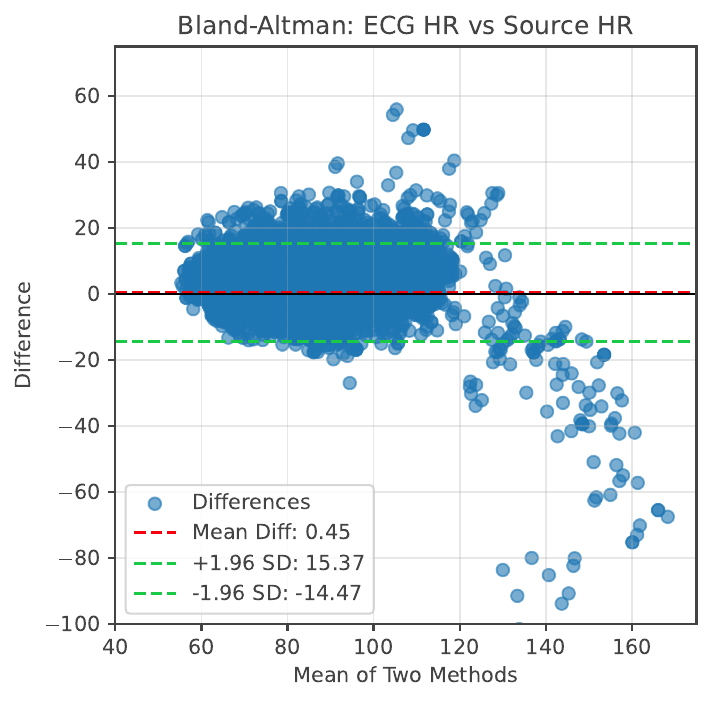}
    \caption{}
    \label{fig:figure5b}
  \end{subfigure}
  \caption{Bland-Altman plots comparing heart rate detection from the original photoplethysmogram (a) and the selected source signal (b) against the reference electrocardiogram.}
  \label{fig:bandaltman}
\end{figure}

The RMSE and correlation values for heart rate (HR), calculated between the original noisy PPG and the reference ECG in the daily routine testing dataset, were $14.4 \pm 10.6$~BPM and $0.407 \pm 0.177$, respectively. For the selected PPG source signal, the values were $4.9 \pm 5.1$~BPM and $0.740 \pm 0.153$. These results, along with the results from ICA, NMF, and BRDAE, are summarized in \autoref{tab:method_comparison}. Bland-Altman plots of the HR errors for both the original PPG and the selected source signal on the same dataset is shown in \autoref{fig:bandaltman}.

\begin{table}[h]
\caption{Comparison of heart rates from various methods against reference ECG HR.}
\centering
\begin{tabular}{lccccc}
\toprule
 & \textbf{None} & \textbf{ICA} & \textbf{NMF} & \textbf{BRDAE} & \textbf{MEAE} \\
\midrule
Correlation & $0.407 \pm 0.177$ & $0.602 \pm 0.214$ & $0.576 \pm 0.181$ & $0.686 \pm 0.147$ & $0.740^* \pm 0.153$ \\
RMSE        & $14.4 \pm 10.6$  & $8.6 \pm 10.4$  & $8.2 \pm 6.0$  & $6.1 \pm 8.2$  & $4.9^* \pm 5.1$ \\
\bottomrule
\end{tabular}
\caption*{*Significant improvement as compared to the original PPG (p<0.01).}
\label{tab:method_comparison}
\end{table}

\section{Discussion}
In this study, the underlying principle of BSS applied to a MEAE network was used to isolate HR from PPG. Testing the trained network on a noisy PPG dataset collected during daily routines of nine subjects, the 1st source separated from the original PPG demonstrated clear HR patterns. As illustrated in \autoref{fig:rpeak}, the BSS approach effectively isolated heart rate-related peaks, even in the presence of significant noise in the original PPG signal. In \autoref{fig:bandaltman}, it can be seen that there is a significant improvement in HR detection from the source signal as compared to the original PPG signal with lower mean error and the standard deviation for the source signal.

Motion artifacts remain one of the primary challenges in obtaining clean PPG measurements \citep{1213624, AllenJohn2007, electronics3020282}, and most studies on PPG denoising focus on mitigating these artifacts. In our previous work, we trained a denoising model using artificially generated noise to expand the range of noise the model could remove. However, motion artifacts, by nature, are inherently random and cannot be fully represented by artificial noise. To address this limitation, the current study aimed to bypass the need for noise-specific denoising by directly extracting the underlying heart rate signal from the PPG, regardless of noise type. By training the MEAE network in a self-supervised manner to separate source signals from the original PPG signal, we avoided reliance on artificially generated noise and achieved superior results compared to our earlier approach.

\subsection{Potential Applications of BSS in Biosignal Processing}
We have demonstrated our BSS method using the example of HR extraction from noisy PPG signals, but the proposed approach has broader applications. As outlined in the Methods section, the MESA dataset was used to train the MEAE network with no data selection, incorporating all available recordings without criteria-based filtering. In supervised neural network training or heuristic-based algorithms, strict data selection and preprocessing often determine performance outcomes. In contrast, the self-supervised approach employed here relies on the inherent distribution of source signals within the original signal, significantly reducing preprocessing requirements.

Moreover, the applicability of the proposed method extends beyond HR detection. Our previous work has demonstrated its capability to extract breathing rate with notable improvements over existing methods \citep{WebsterMatthew2024} and similar results can be seen on MESA results in \autoref{fig:rpeak}, where the 4th and 6th source signals seem to indicate breathing patterns, underscoring the method's versatility. By modifying the model’s hyperparameters, it is possible to extract different types of source information from different types of biosignal, further enhancing its utility.

\begin{figure}[htbp]
  \centering
  \includegraphics[width=\linewidth]{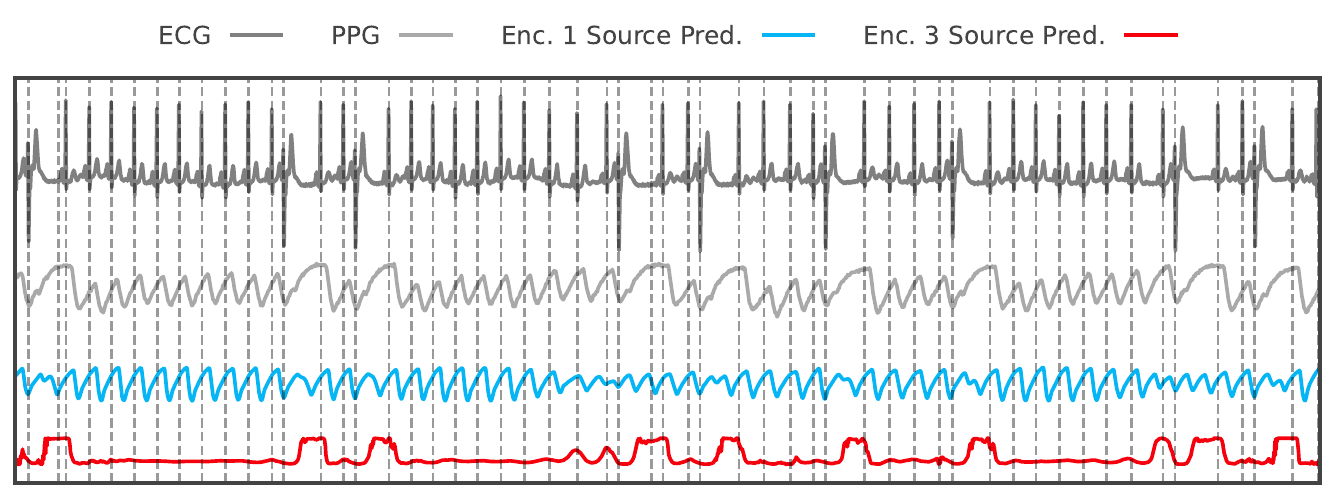}
  \caption{A segment from the MESA dataset with arrythmia. The 1st source signal shown (Enc. 1) captures most of the heart beats present in the electrocardiogram, while larger gaps in the heart peaks are aligned with the peaks shown in the 2nd source signal (Enc. 3).
}
  \label{fig:arryth}
\end{figure}

In addition to HR extract from PPG, an interesting finding emerged when analyzing other HR-related information in the ECG data. In \autoref{fig:arryth}, the 2nd source signal (encoder 1) captured most of the heartbeats within the source ECG segment, while the 4th (encoder 3) source exhibited peaks at specific time intervals. Upon closer inspection of the corresponding ECG segment, arrhythmia and irregular heartbeats can be seen around the 4th source signal’s peaks. Although the model was not explicitly trained to detect such abnormalities, it seems to have isolated arrhythmia (or some correlated signal) as a distinct source signal for this particular PPG segment.

Further analysis revealed that the model did not identify all instances of arrhythmia within the dataset, indicating that detecting such irregularities requires additional investigation. Nonetheless, this result highlights the potential of the proposed BSS method for broader biosignal processing applications, including the identification of pathological events.

\subsection{Future Works to Improve BSS in Biosignals}
The current MEAE architecture is not well suited for determined and overdetermined mixtures due to the fact that each channel in the observation results in its own set of sources equal to the number of encoders. This fact limits the type of data that can be used to single-channel underdetermined mixtures. For signals such as multi-channel ECG or EEG, determining a method to incorporate the source contributions within each channel may be a viable solution in future works.

Further, the model does not seem to generalize to the case in which no sources are present within the input mixture. This can be due to severe distortions in the mixture that destroy all relevant or necessary information for source reconstruction. Adding a mechanism that can detect these scenarios would help to prevent the model from hallucinating sources where none are actually present. Ideally this mechanism for detecting the presence of sources would also be fully self-supervised, but supervised approaches with synthetic signals and distortions may also be effective.

Future works should also explore the design and hyperparameter optimization of the sparse mixing loss. Currently, different decay values for the sparse mixing regularization can produce very different results as different values make different assumptions about the mixing system. Efficiently finding the appropriate value or set of values for a given problem is an important next step in improving MEAE for BSS. Additionally, it may be valid to apply different amounts of decay to different layers or even sets of mixing weights within a given layer.

\subsection{Limitations}
Although the proposed method outperformed our previous method using BRDAE and artificial noise augmentation, the performance for extracting HR-related source signal can be improved. Since the performance of source signal separation in the proposed method depends entirely on the distribution of the training data, any noise sources or distortions present in the testing dataset but absent in the training data can negatively impact the results. However, we could not identify an open dataset large enough to train the MEAE on a sufficient variety of noise profiles associated with daily activities. This limitation highlights the need for future research to address the challenges of training data distribution. A potential solution could involve training the MEAE on a large dataset with limited variation and fine-tuning it on a smaller dataset with greater diversity of features.

Another limitation of the proposed method is the absence of an explicit mechanism for halting training. The reconstruction loss converges relatively quickly to a stable value, but the outputs associated with each individual encoders (i.e. the source signals) vary across training epochs. This variation arises because there is no unique solution that minimizes the loss; different encoders may capture different source signals while achieving the same overall loss. Consequently, the outputs of the encoders can differ significantly at different training stages. To address this issue, we introduced a manual evaluation step. After each training epoch, we input the testing PPG data from the nine subjects into the MEAE and calculated the errors in HR estimation for each encoder channel compared to the reference ECG. These errors were tracked throughout the training process, and the best-performing epoch was selected for result presentation. In future work, we aim to develop a more robust loss function to constrain the distribution of outputs across the encoders. One potential approach is to incorporate a loss term based on the frequency characteristics of each encoder’s output, ensuring that the outputs are aligned with the expected source signal distributions. This enhancement could further improve the reliability and generalizability of the proposed method.

\section{Code availability statement}
Our implementation of the experiments described in this paper is available on GitHub at the following link: \href{https://github.com/webstah/meae-heart-rate-extraction-from-noisy-ppg}{\texttt{github.com/webstah/meae-heart-rate-extraction-from-noisy-ppg}}

The Github repository containing the implementation of the prior study on MEAE \citep{WebsterMatthew2024} can be found here:
\href{https://github.com/webstah/self-supervised-bss-via-multi-encoder-ae}{\texttt{github.com/webstah/self-supervised-bss-via-multi-encoder-ae}}

\section{Ethics statement}
\paragraph{MESA dataset}The MESA dataset is available from the National Sleep Research (\href{https://www.sleepdata.org}{\texttt{www.sleepdata.org}}) following an application process for access.
\paragraph{9-subject noisy photoplethysmogram dataset}This study was conducted in accordance with the ethical standards of the institutional and/or national research committee and with the 1964 Helsinki declaration and its later amendments or comparable ethical standards. All subjects provided informed consent and the study was approved by the Institutional Review Board of Seoul National University Hospital. 

\section{Funding}
This work was supported by Chungnam National University Hospital Research Fund, 2023.

\section{Declaration of generative AI and AI-assisted technologies in the writing process}
During the preparation of this work the authors used ChatGPT in order to clarify some of the obscure wording in the manuscript. After using this tool/service, the authors reviewed and edited the content as needed and takes full responsibility for the content of the published article.

\bibliographystyle{unsrtnat}
\bibliography{references} 

\end{document}